\documentclass[11pt]{article}

\usepackage[final]{acl}

\usepackage{xcolor}
\usepackage{times}
\usepackage{latexsym}

\usepackage[T1]{fontenc}

\usepackage[utf8]{inputenc}

\usepackage{microtype}

\usepackage{inconsolata}

\usepackage{graphicx}

\usepackage{amsmath}
\usepackage{amssymb}
\usepackage{booktabs}
\usepackage{multirow}

\newcommand{\imp}[1]{{\scriptsize\color{teal}(+#1)}}
\newcommand{\dec}[1]{{\scriptsize\color{red}(-#1)}}

\usepackage{pifont}        

\usepackage{array}         
\usepackage{makecell}      
\usepackage{subcaption}

\usepackage{tcolorbox}
\tcbuselibrary{skins} 
\usepackage{enumitem}

\newtcolorbox{promptbox}[2][]{
  colback=gray!5,        
  colframe=black!70,     
  title=\textbf{#2},     
  fonttitle=\bfseries\small,
  fontupper=\footnotesize\ttfamily, 
  left=3pt, right=3pt, top=3pt, bottom=3pt, 
  boxrule=0.8pt,
  sharp corners,         
  #1
}

\title{Beyond Quantity: Trajectory Diversity Scaling for Code Agents}

\author{
 \textbf{Guhong Chen\textsuperscript{1,2}}\thanks{Equal contribution.}\thanks{Work was done when interned at Qoder.},
 \textbf{Chenghao Sun\textsuperscript{1}}\footnotemark[1],
 \textbf{Cheng Fu\textsuperscript{2}}\footnotemark[1],
 \textbf{Qiyao Wang\textsuperscript{1,2}},
 \textbf{Zhihong Huang\textsuperscript{1,2}},
\\
 \textbf{Chaopeng Wei\textsuperscript{1}},
 \textbf{Guangxu Chen\textsuperscript{1}},
 \textbf{Feiteng Fang\textsuperscript{1}},
 \textbf{Ahmadreza Argha\textsuperscript{4}},
 \textbf{Bing Zhao\textsuperscript{3}},
\\
 \textbf{Xander Xu\textsuperscript{3}},
 \textbf{Qi Han\textsuperscript{3}},
 \textbf{Hamid Alinejad-Rokny\textsuperscript{4}},
 \textbf{Qiang Qu\textsuperscript{1}},
 \textbf{Binhua Li\textsuperscript{2}},
\\
 \textbf{Shiwen Ni\textsuperscript{1,5}}\thanks{Corresponding authors: Shiwen Ni, Min Yang, and Hu Wei.},
 \textbf{Min Yang\textsuperscript{1,5}}\footnotemark[3],
 \textbf{Hu Wei\textsuperscript{3}}\footnotemark[3],
 \textbf{Yongbin Li\textsuperscript{2}},
 \textbf{Yu Ding\textsuperscript{2}},
\\
 \textsuperscript{1}SIAT, CAS \quad
 \textsuperscript{3}Alibaba Group \quad
 \textsuperscript{4}UNSW Sydney \quad
 \textsuperscript{5}SUAT\\
 \textsuperscript{2}Qoder \raisebox{-0.15ex}{\includegraphics[height=2.0ex]{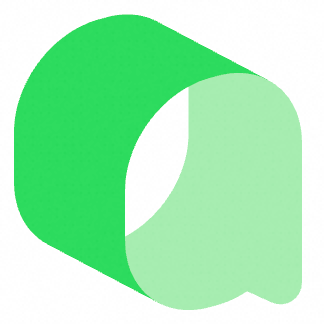}}
\\
\texttt{\{gh.chen2, sw.ni\}@siat.ac.cn}
}

\begin{document}
\maketitle

\begin{abstract}
As code large language models (LLMs) evolve into tool-interactive agents via the Model Context Protocol (MCP), their generalization is increasingly limited by low-quality synthetic data and the diminishing returns of quantity scaling; moreover, quantity-centric scaling exhibits an early bottleneck that underutilizes trajectory data. We propose \textbf{TDScaling}, a \textbf{T}rajectory \textbf{D}iversity \textbf{Scaling}-based data synthesis framework for code agents that scales performance through \emph{diversity} rather than raw volume. Moreover, TDScaling is more data-efficient: under a fixed training budget, increasing trajectory \emph{diversity} yields larger gains than adding more trajectories, improving the performance--cost trade-off for agent training. TDScaling integrates four innovations: (1) a Business Cluster mechanism that captures real-service logical dependencies; (2) a Blueprint-driven multi-agent paradigm that enforces trajectory coherence; (3) an adaptive evolution mechanism that steers synthesis toward long-tail scenarios using Domain Entropy, Reasoning Mode Entropy, and Cumulative Action Complexity to prevent mode collapse; and (4) a sandboxed code tool that mitigates catastrophic forgetting of intrinsic coding capabilities. Experiments on general tool-use benchmarks (BFCL, $\tau^2$-Bench) and code agent tasks (RebenchT, CodeCI, BIRD) demonstrate a win--win outcome: TDScaling improves both tool-use generalization and inherent coding proficiency. Crucially, we show that trajectory diversity scaling attains a substantially higher performance ceiling than quantity scaling, establishing a resource-efficient paradigm for training robust code agents under data bottlenecks. \footnote{We plan to open-source all components that can be publicly released, subject to licensing, privacy, and internal policy constraints.}
\end{abstract}

\section{Introduction}

\begin{figure}[t]
    \centering
    \includegraphics[width=1\linewidth]{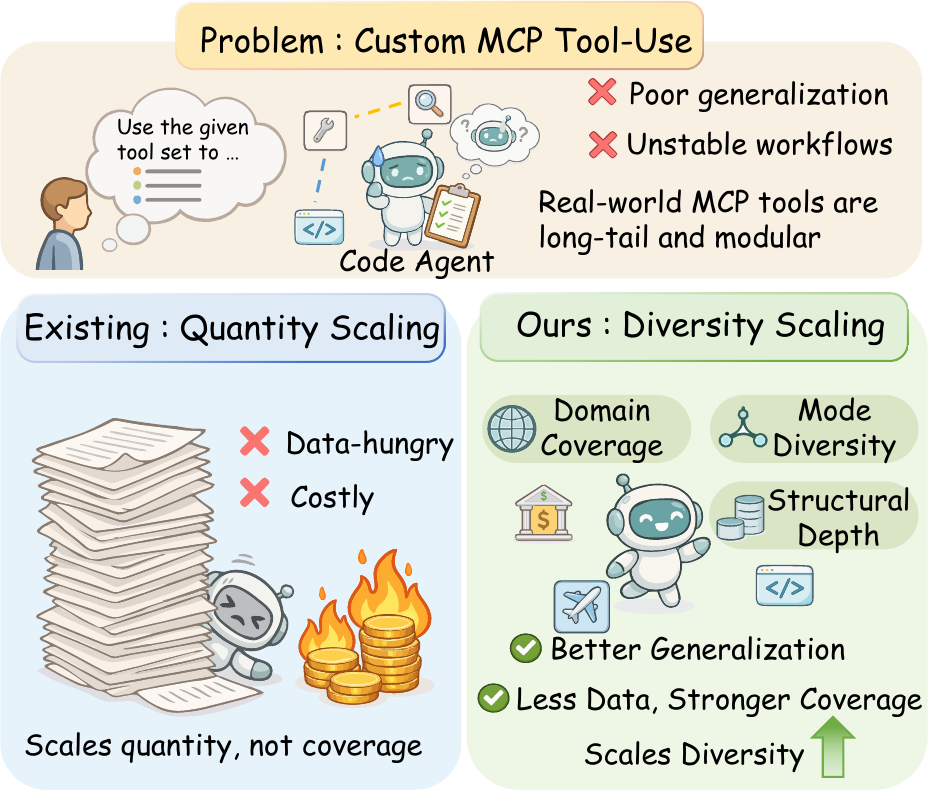}
   \caption{\textbf{Contrast between Existing Quantity Scaling and our Diversity Scaling.} Instead of relying on costly, data-hungry expansion, our approach optimizes for trajectory diversity. This strategy addresses the poor generalization of Code Agents in custom MCP environments, achieving stronger robustness with a smaller, high-quality dataset.}
    \label{fig:diversity-scale-compare}
\end{figure}

Software engineering is being reshaped by tool-interactive agents. With protocols such as the Model Context Protocol (MCP)~\cite{anthropic2024mcp}, code large language models (LLMs) are moving beyond static generation toward agents that can invoke, compose, and debug external utilities. In practice, strong developers succeed by coordinating a heterogeneous tool ecosystem; likewise, the competitiveness of next-generation coding agents depends not only on producing syntactically correct code, but on selecting and composing tools under evolving specifications and failures~\cite{yao2022react, schick2023toolformer}. 

However, current training paradigms face a bottleneck in generalizing to diverse or dynamically registered tools. While models such as Qwen-Coder~\cite{hui2024qwen2} are strong at algorithmic logic, their performance degrades on unfamiliar tool interfaces and interaction patterns. In practice, these failures concentrate in long-horizon interactions---tool selection, composition, and error recovery---where agents must interpret new specifications and adapt actions on the fly. This gap suggests that many models rely on parametric recall of known APIs~\cite{patil2024gorilla, qin2023toolllm} instead of robust in-context reasoning over new tool specifications~\cite{li-etal-2023-api}.

A common response is to scale synthetic data quantity, but quantity-centric scaling often yields diminishing returns. Existing datasets~\cite{qin2023toolllm, chen-etal-2025-acebench} can be domain-homogeneous and dominated by simple, repetitive interactions. Increasing the volume of such low-entropy trajectories does not adequately cover long-tail behaviors (e.g., nested tool calls, exception handling, and recovery), leading to an early performance ceiling. This matters because MCP-style environments continually introduce new tools and evolving schemas, so agents that cannot generalize beyond seen APIs remain brittle and difficult to deploy. 

To overcome this limitation, we propose \textbf{TDScaling} (Trajectory Diversity Scaling), which shifts synthetic scaling from \emph{quantity} to \emph{diversity} to better exploit trajectory data.
As illustrated in Figure~\ref{fig:diversity-scale-compare}, we advocate for a shift to diversity Scaling, optimizing for domain coverage and structural depth to achieve robust generalization with significantly higher data efficiency.

First, to address tool coverage, we introduce a Business Cluster-based sampling mechanism. Instead of random sampling that produces redundant and weakly related APIs, we organize the MCP ecosystem into coherent semantic clusters. This design increases semantic coverage under limited budgets and better reflects real-service logical dependencies, yielding representative toolsets that support diversity-oriented synthesis and downstream scaling-law analysis.

Second, to drive synthesis toward challenging behaviors, we introduce an adaptive evolution mechanism guided by quantifiable metrics. Rather than relying on rigid templates, our system promotes trajectory diversity by optimizing Reasoning Mode Entropy and Cumulative Action Complexity, dynamically identifying and filling distributional gaps. This process steers generation toward under-explored, high-complexity regions such as multi-step composition and error recovery. We further integrate a sandboxed code tool as a regularizer: combining standard tool invocations with programmatic reasoning strengthens verification and mitigates catastrophic forgetting of intrinsic coding ability that can arise during tool tuning.

Experiments on general tool-use benchmarks (BFCL, $\tau^2$-Bench) and agentic coding tasks (RebenchT, CodeCI, BIRD) show a win--win effect: TDScaling improves both tool-use generalization and inherent coding proficiency. In particular, TDScaling enables Qwen3-Coder-30B-A3B to reach performance comparable to 480B-scale models on these evaluations. Moreover, our analysis shows that diversity scaling achieves a higher performance ceiling than quantity scaling, offering a more resource-efficient paradigm for training robust code agents.

Our main contributions are as follows:
\begin{itemize}
\item We present \textbf{TDScaling}, a diversity-first synthesis paradigm for code agents, and empirically establish that \emph{diversity} scaling surpasses \emph{quantity} scaling in attainable performance ceiling.
\item We propose \textbf{Business Cluster} sampling that captures real-service logical dependencies, yielding high-coverage toolsets with substantially reduced redundancy.
\item We develop an \textbf{entropy/complexity-guided evolution} strategy with a \textbf{sandboxed code-tool regularizer} that targets long-tail interaction patterns (e.g., composition and error recovery) while mitigating catastrophic forgetting of coding skills.
\end{itemize}

\section{Related Work}
\subsection{Generalizing Tool-Use Capabilities in Code Agents}
The growing complexity of software development has increased demand for automated code generation and intelligent programming assistants. Large language models (LLMs) have progressed from static code completion to supporting more autonomous software engineering workflows. In this setting, specialized open-weight models such as StarCoder \citep{lozhkov2024starcoder}, DeepSeek-Coder \citep{zhu2024deepseek}, and the Qwen-Coder series \citep{hui2024qwen2} show strong proficiency in programming syntax and logic, providing a solid foundation for building code agents.

Beyond code synthesis, expanding an LLM's action space through external tools---often referred to as tool learning---is central to enabling more capable agents \citep{qu2025tool}. Recent models such as DeepSeek-V3.2 \citep{liu2025deepseek} report substantial gains by strengthening tool-use capability.

Within code agents, tool integration has largely focused on the coding phase. As surveyed by \citet{dong2025survey}, existing systems extend beyond code execution to incorporate retrieval of API documentation \citep{ding-etal-2025-toolcoder}, static analyzers for execution feedback \citep{zhang-etal-2023-self}, and dependency resolution \citep{zhang-etal-2024-codeagent}. However, these tools are typically specialized for programming-centric workflows and do not directly address general tool-use over heterogeneous external services. As platforms such as Cursor adopt MCP to connect diverse services \citep{cursor2025mcp}, code LLMs must generalize from narrow coding utilities to tool ecosystems with evolving interfaces and state. Our work targets this gap by improving generalizable tool-use capability for code agents operating in MCP environments.

\subsection{Synthesizing and Verifying Tool-Use Trajectories}
The limited availability of high-quality and diverse tool-use trajectories has motivated multi-agent frameworks for synthetic data generation. Recent approaches \citep{mitra2024agentinstruct, tang-etal-2025-synthesizing, liu2024toolace} coordinate multiple LLM agents to produce multi-turn instruction--response data, scaling training corpora beyond what manual annotation can support.

A complementary trend emphasizes verification to improve synthesis reliability. APIGen \citep{prabhakar2025apigen} introduces a Blueprint-driven mechanism and filters trajectories through multi-stage validation that combines execution checks with semantic review. DeepSeek-V3.2 \citep{liu2025deepseek} further leverages executable environments and programmatic rewards at scale, and TOUCAN \citep{xu2025toucan} grounds synthesis by interacting with real-world MCP servers. To reduce the engineering burden of maintaining execution backends, Simia \citep{li2025simulating} proposes using reasoning models to simulate environment responses. Despite these advances, execution-centric methods remain constrained by environment availability, while pure simulation can break the stateful dependencies of real services. In contrast, our framework preserves semantic coherence by organizing tools into Business Clusters and maintains logical coupling through a Blueprint-driven mechanism, enabling scalable synthesis of high-complexity, logically consistent trajectories without relying on heavy execution infrastructures.

\begin{figure*}[t]
    \centering
    \includegraphics[width=\linewidth]{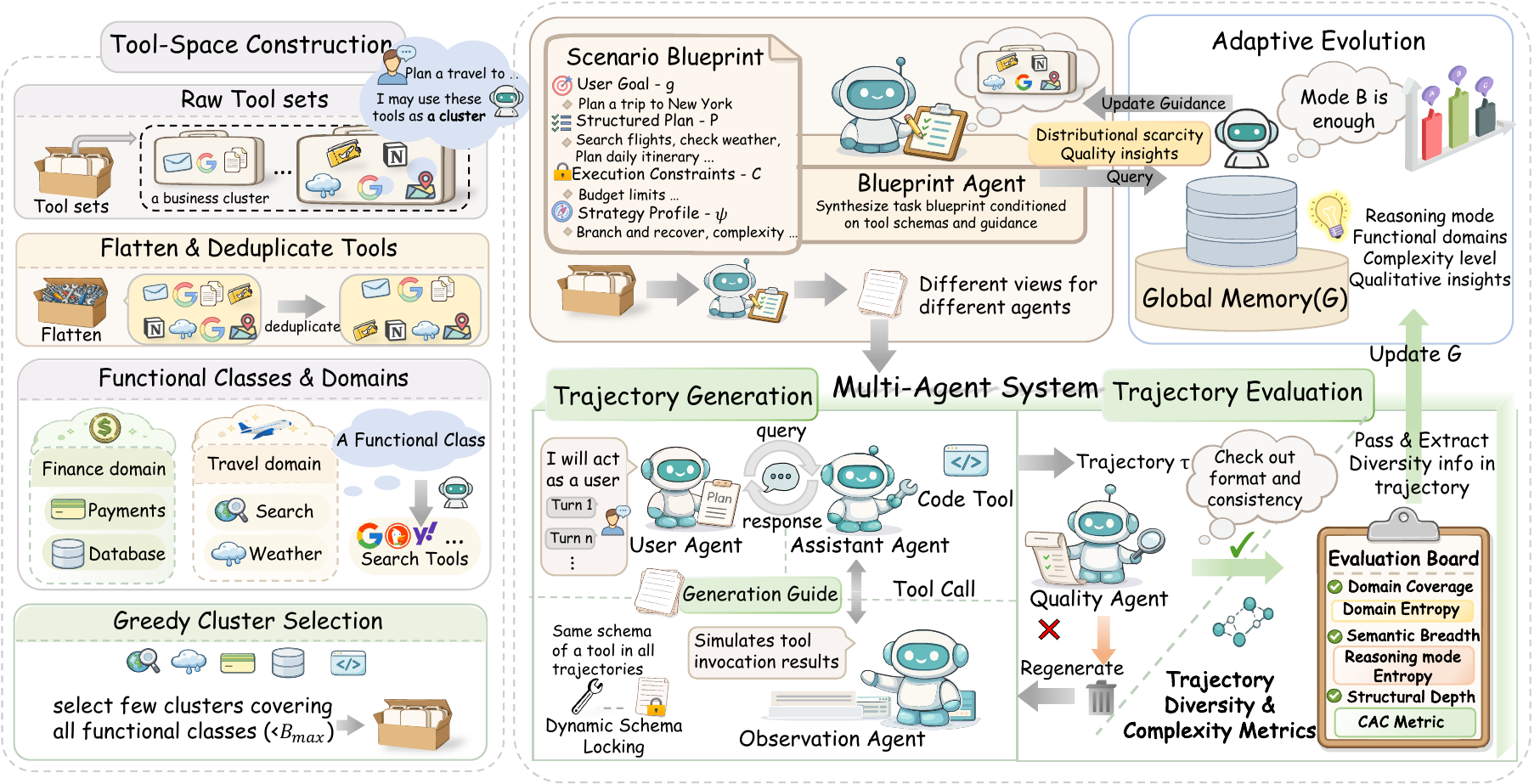}
    \caption{\textbf{Overview of the TDScaling framework.} The pipeline has four stages: (1) Tool-Space Construction: Raw MCP definitions are organized into Business Clusters and filtered via greedy selection to maximize functional coverage. (2) Blueprint Synthesis: Conditioned on the selected toolset and Global Memory scarcity, a Blueprint Agent generates a Scenario Blueprint with goals, plans, constraints, and strategies. (3) Multi-Agent Execution: User, Assistant, and Observation agents generate trajectories, dynamically invoking a Code Tool for programmatic reasoning, while a Quality Agent enforces format adherence and logical consistency. (4) Adaptive Evolution: Validated trajectories are scored for diversity and complexity; successful traces update Global Memory, which adjusts strategy profiles to steer generation toward under-explored reasoning modes and higher-complexity regions.}

    \label{fig:pipeline}
\end{figure*}

\section{TDScaling Framework}

\subsection{Tool-Space Construction via Business Clusters}
\label{sec:problem_setting}

\paragraph{Leveraging MCP Servers as Business Clusters.}
Our framework leverages a large-scale collection of real-world MCP tool definitions. Distinct from previous works that flatten tool repositories into isolated API endpoints, we strictly respect the native modularity of the Model Context Protocol by treating each MCP Server as a Business Cluster ($\mathcal{B}$). Preserving this structure allows the model to learn coherent, dependency-aware workflows inherent in real-world services.

\paragraph{Business Cluster-based Sampling.}
We formulate the dataset construction as a Maximum Coverage Problem to select a subset of clusters $\mathcal{S} \subseteq \mathcal{B}$ under a budget constraint $B_{\max}$. This strategy prioritizes semantic breadth over naive random sampling:
\begin{equation}
  \max_{\mathcal{S}} \Bigl\lvert \bigcup_{B_i \in \mathcal{S}} F(B_i) \Bigr\rvert
  \quad \text{s.t.} \quad |\mathcal{S}| \le B_{\max}
  \label{eq:max_coverage}
\end{equation}
where $F(B_i)$ denotes the set of unique functional classes in cluster $B_i$. A greedy approximation solves this problem, and an intra-cluster refinement prunes redundant tools. Detailed construction steps and are provided in Appendix~\ref{app:implementation_details}.

\subsection{Scenario Blueprinting and Multi-Agent Synthesis}

To ensure synthesized trajectories possess logical depth, we employ a Blueprint-then-Execute paradigm. This approach mitigates hallucination risks by anchoring interactions to a pre-computed logic topology.

\paragraph{Scenario Blueprint.}
For a selected cluster $B_i$, the BlueprintAgent generates a Scenario Blueprint $\mathcal{S}_{bp} = (g_i, P_i, C_i, \Psi_i)$, where $g_i$ is the user goal, $P_i$  the execution plan, $C_i$ specifies constraints, and $\Psi_i$ a Strategy Profile. The Strategy Profile guides synthesis style (e.g., prioritizing nested tool calls) via  feedback from global memory.

\paragraph{Multi-Agent Execution \& Consistency.}
Trajectories are synthesized through a collaborative role-play loop.
The \textbf{UserAgent} executes $P_i$, while the \textbf{AssistantAgent} generates reasoning traces and tool calls.
To ensure simulation fidelity, the \textbf{ObservationAgent} employs a Dynamic Schema Locking mechanism.
In synthetic environments, a common failure mode is ``structural hallucination,'' where the simulator returns inconsistent JSON schemas for the same tool across turns. To counteract this, our agent caches the output schema generated in the first turn and strictly enforces structural adherence in all subsequent calls ($T+1, \dots, N$). This constraint forces the model to learn stable API contracts rather than adapting to shifting simulator artifacts.
Finally, the \textbf{QualityAgent} ensures Context-Response Consistency, and validated trajectories update global memory to refine $\Psi_i$. System prompts are detailed in Appendix~\ref{app:prompts_system} (Figures~\ref{fig:prompt_foundation}--\ref{fig:prompt_environment}).

\subsection{Code Tool as a Regularizer}

We integrate a sandboxed Python Code Tool to empower the agent with complex data processing capabilities while preserving its coding proficiency.
To ensure the model prioritizes standard APIs, we employ a General-Tool-First principle: the code tool is dynamically injected only when the BlueprintAgent identifies that the task requires computational logic unsolvable by standard functional tools. As illustrated in Figure~\ref{fig:code_tool_appendix}, while standard agents struggle with multi-criteria sorting (e.g., specific impact and recency weights), the injected Code Tool ensures 100\% execution accuracy through programmatic logic.

Integrating the code tool serves dual strategic objectives. First, it enables Program-of-Thought reasoning, allowing the model to offload internal logic to a deterministic interpreter. Second, and critically, it acts as a regularizer against catastrophic forgetting. By interleaving substantive code generation within tool-use trajectories, we ensure the training distribution aligns with the model's pre-training priors, effectively reversing the negative transfer often observed in API-centric fine-tuning.

\subsection{Adaptive Evolution via Diversity Metrics}
\label{sec:metric_definitions}
To prevent reasoning pattern convergence and drive iterative quality improvement, we employ an adaptive evolution mechanism backed by a global memory $G$. This mechanism is guided by three quantifiable dimensions of diversity (formal mathematical definitions are detailed in Appendix~\ref{app:metric_definitions}).

\paragraph{Domain Coverage: Business Cluster Entropy.}
Complementary to reasoning styles, we first measure the semantic span of the tool ecosystem. Mapping each synthesized trajectory $\tau$ to its primary Business Cluster $B_k$ (as defined in Sec.~\ref{sec:problem_setting}), we calculate the normalized domain probability distribution $p(B_k)$. We define the Domain Entropy as:
\begin{equation}
  H_{\text{dom}} = - \sum_{B_k \in \mathcal{B}} p(B_k) \log p(B_k)
  \label{eq:domain_entropy}
\end{equation}
Maximizing $H_{\text{dom}}$ prevents collapse into a few dominant tool categories, guiding the system to fill gaps in the tool-use landscape.

\paragraph{Semantic Breadth: Reasoning Mode Entropy.}
Standard generation tends to gravitate towards low-effort reasoning paths. Unlike rule-based systems, we adopt a data-driven approach. For each trajectory, the QualityAgent analyzes the interaction flow and assigns a reasoning label $m$. 
Crucially, we do not restrict $m$ to a predefined list; instead, the agent is encouraged to dynamically identify and tag novel reasoning patterns (e.g., \textit{Hypothesis-Testing}, \textit{Recursive-Correction}) that emerge during the exploration of new tool clusters.
We quantify breadth using Shannon entropy over the empirical frequency of these dynamically discovered modes:
\begin{equation}
  H_{\text{mode}} = - \sum_{m \in M} p(m) \log p(m)
  \label{eq:mode_entropy}
\end{equation}
Increasing $H_{\text{mode}}$ indicates the successful injection of diverse reasoning structures beyond trivial model bias.

\paragraph{Structural Depth: Cumulative Action Complexity.}

We measure the intrinsic execution difficulty via Cumulative Action Complexity (CAC). We decompose the cognitive load of an action $a_i$ into the product of lateral tool selection costs and hierarchical argument instantiation costs:
\begin{equation}
\mathcal{C}(a_i)
= \mathcal{C}_{\text{switch}}(t_i \mid t_{i-1})
\cdot
\mathcal{C}_{\text{depth}}(\theta_i \mid \mathcal{H}_i)
\label{eq:action_load}
\end{equation}
The switching cost $\mathcal{C}_{\text{switch}}$ models the cognitive complexity of shifting functional contexts. Let $\phi(t)$ denote the tool-to-domain mapping, the cost is then formulated as:
\begin{equation}
\small
\mathcal{C}_{\text{switch}}(t_i \mid t_{i-1}) = 
\begin{cases} 
\mu_{\text{base}} & i=1 \\
\mu_{\text{base}} + \delta \cdot \mathbb{I}[\phi(t_i) \neq \phi(t_{i-1})] & i > 1
\end{cases}
\label{eq:dswitch}
\end{equation}
The depth cost $\mathcal{C}_{\text{depth}}$ measures the information lineage required for argument instantiation. We estimate a dependency level $y(p)$ for each parameter (Instruction-Grounded, Local-Context, or Global-Context; detailed definitions and weights are provided in Appendix~\ref{app:metric_definitions} and define the cost as the bottleneck weight:

\begin{equation}
\mathcal{C}_{\text{depth}}(\theta_i \mid \mathcal{H}_i) = \max_{p \in \theta_i} \omega_{y(p)}
\label{eq:depth_cost}
\end{equation}

Driven by this tuple $(H_{\text{dom}}, H_{\text{mode}}, \text{CAC})$, the BlueprintAgent queries $G$ to identify distributional gaps. The system then dynamically steers synthesis toward under-explored scenarios to maximize data potential.
\section{Experiments}
\label{sec:experiments}

\subsection{Experimental Setup}

\noindent\textbf{Models and Baselines.}
We adopted the Qwen3-Coder family (Qwen3-Coder-30B-A3B-Instruct) as the primary backbone to evaluate agentic coding capability. We also evaluated the general-purpose Qwen3-30B-A3B-Instruct to assess the universality of TDScaling, i.e., whether it improved tool-use beyond specialized coding models.
We compared against strong proprietary models and leading open-source baselines.
For tool-learning methods (APIGen-MT, TOUCAN, Simia), we evaluated checkpoints trained on 5,000 samples, which reflected the maximum common data availability across these open-source projects and enabled a fair comparison at their accessible limit.
For TDScaling, we used a dual-scale protocol: we first evaluated with a minimal set of \textbf{500 samples} to stress-test data efficiency, and then scaled to \textbf{5,000 samples} to compare performance ceilings under matched data budgets.

\noindent\textbf{Dataset and Training.}
The source environment consisted of 30,000 raw MCP-compliant tool definitions.
From this pool, we applied the greedy sampling strategy (Sec.~\ref{sec:problem_setting}) to select 6,944 high-quality Business Clusters, preserving real-world logical dependencies.
During clustering, we persisted functional domain mappings to support the computation of quantitative metrics (e.g., Entropy and Complexity) during evaluation.
Using Qwen3-Max as the teacher, we synthesized complex tool-use trajectories from these clusters via the Blueprint-driven evolutionary framework.
We fine-tuned models directly on the synthesized trajectories without mixing in other general instruction data.
All models were trained with Megatron-LM in BF16 precision; detailed hyperparameters were reported in Appendix~\ref{app:implementation_details}.

\noindent\textbf{Evaluation Benchmarks.}
We benchmarked performance along two dimensions.
For \textit{General Tool Use}, we used:
(1) \textbf{BFCL}~\cite{patil2024gorilla}: We used the Augmented Multi-Turn subset to evaluate stateful reasoning under complex conditions, including missing parameters, missing functions, and long-context dependencies, which required clarification and robust decision-making beyond direct execution.
(2) \textbf{$\tau^2$-Bench}~\cite{barres2025tau}: A dual-control Dec-POMDP environment that tested dynamic coordination with active users who modified a shared world state.

For \textit{Coding and Agentic Tasks}, we used:
(1) \textbf{SWE-rebench (RebenchT)}~\cite{badertdinov2025swe}: An interactive benchmark derived from real-world GitHub issues to assess repository navigation and engineering adaptation.
(2) \textbf{CodeCI (LiveCodeBench)}~\cite{jain2024livecodebench}: Adapted from release v6 with a custom interpreter to evaluate end-to-end skills such as self-repair and test prediction.
(3) \textbf{BIRD}~\cite{li2023can}: A large-scale (33.4 GB) text-to-SQL benchmark with dirty values that required complex semantic parsing beyond template-based generation.

\subsection{Main Results}

\paragraph{General Tool-Use Capabilities.}
Table~\ref{tab:tooluse_main} summarizes performance on the general tool-use benchmarks. Our method consistently outperformed the open-source baselines.
A key finding was that Qwen3-Coder-30B-A3B fine-tuned with TDScaling reached 36.66\% on the challenging BFCL Multi-turn benchmark with only 500 samples, exceeding the much larger Qwen3-Coder-480B-A35B-Instruct baseline (35.91\%). This result supported the value of evolutionary distillation: a smaller model trained on high-diversity evolved trajectories could outperform a substantially larger model trained on standard data.
With the same 500-sample budget, TDScaling achieved an average score of 48.99, surpassing fully trained baselines such as APIGen-MT (42.81) and Simia (45.29). When scaled to 5,000 samples, TDScaling reached 40.44\% on BFCL, indicating a higher performance ceiling under aligned data budgets.

\begin{table*}[t]
\centering
\small
\setlength{\tabcolsep}{6.4pt}
\begin{tabular}{lccccc}
\toprule
\textbf{Model} & \textbf{BFCL} & \textbf{TAU-AIR} & \textbf{TAU-RET} & \textbf{TAU-TEL} & \textbf{Average} \\
& \textit{(Multi-turn)} & & & & \\
\midrule
\multicolumn{6}{l}{\textit{Proprietary Models}} \\
GPT-5 & 43.75 & 58.00 & 77.20 & 95.80 & 68.69 \\
GPT-4.1 & 38.88 & 56.00 & 74.00 & 34.00 & 50.72 \\
Claude-Sonnet-4 & 54.75 & 67.50 & 54.00 & 47.40 & 55.91 \\
Gemini-2.5-pro & 29.25 & 67.50 & 56.00 & 27.20 & 44.99 \\
\midrule
\multicolumn{6}{l}{\textit{Open-Source Foundation Models}} \\
DeepSeek-V3.2 & 44.88 & 63.80 & 74.12 & 96.20 & 69.75 \\
Qwen3-Coder-480B-A35B-Instruct & 35.91 & 41.00 & 63.82 & 66.67 & 51.85 \\
\midrule
\multicolumn{6}{l}{\textit{Tool-Learning Methods (5k Samples)}} \\
APIGen-MT ~\cite{prabhakar2025apigen} & 27.25 & 33.00 & 60.75 & 50.22 & 42.81 \\
TOUCAN ~\cite{xu2025toucan} & 37.03 & 33.50 & 56.36 & 56.80 & 45.92 \\
Simia ~\cite{li2025simulating} & 23.22 & 52.00 & 58.77 & 47.15 & 45.29 \\
\midrule
\multicolumn{6}{l}{\textit{TDScaling Implementation (Ours)}} \\
Qwen3-30B-A3B-Instruct & 33.22 & 30.50 & 53.29 & 21.93 & 34.74 \\
\textbf{TDScaling (500 Samples)} & 36.63 \imp{3.41} & 39.00 \imp{8.50} & 58.55 \imp{5.26} & 31.80 \imp{9.87} & 41.50 \imp{6.76} \\
\addlinespace[4pt] 
Qwen3-Coder-30B-A3B-Instruct & 29.41 & 36.50 & 58.55 & 41.45 & 41.48 \\
\textbf{TDScaling (500 Samples)} & 36.66 \imp{7.25} & 40.00 \imp{3.50} & 63.38 \imp{4.83} & 55.92 \imp{14.47} & 48.99 \imp{7.51} \\
\textbf{TDScaling (5000 Samples)} & \textbf{40.44} \imp{11.03} & \textbf{44.00} \imp{7.50} & \textbf{64.69} \imp{6.14} & \textbf{60.75} \imp{19.30} & \textbf{52.47} \imp{10.99} \\
\bottomrule
\end{tabular}
\caption{
Performance on general tool-use benchmarks.
We compared TDScaling against proprietary models and open-source foundation models.
\textbf{Average} is computed over BFCL and the three $\tau^2$-Bench domains.
The absolute improvements (\textcolor{teal}{+gain}) in the bottom block show that TDScaling delivered substantial gains with minimal data (500 samples) and continued to scale to 5,000 samples.
}

\label{tab:tooluse_main}
\end{table*}

\paragraph{Coding Agent and Programmatic Reasoning.}
Table~\ref{tab:coding_main} reports results on the agentic coding tasks.
A common failure mode in tool tuning is negative transfer: optimizing for API invocation can erode general reasoning and coding ability. This trend appeared in baselines such as APIGen-MT and Simia, which underperformed the base model (30.99\% Overall).
In contrast, TDScaling produced a positive gain (+4.00\% Overall), reaching 34.99\%.
By integrating the Code Tool to mitigate catastrophic forgetting of intrinsic coding capabilities, TDScaling aligned tool-use training with the model's pre-training priors and encouraged programmatic reasoning, benefiting both general tool use and coding-centric tasks.

\begin{table*}[t]
\centering
\small
\setlength{\tabcolsep}{7pt} 
\begin{tabular}{lccccc}
\toprule
\multirow{2}{*}{\textbf{Model}} & \multicolumn{2}{c}{\textbf{RebenchT}} & \textbf{CodeCI} & \textbf{Bird} & \textbf{Overall} \\
\cmidrule(lr){2-3} 
 & \textbf{OH-p@1} & \textbf{Qod-p@1} & \textbf{avg@2} & \textbf{p@1} & \textit{(Avg)} \\
\midrule
\multicolumn{6}{l}{\textit{Baseline}} \\
Qwen3-Coder-30B-A3B-Instruct & 31.21 & 15.84 & 35.43 & 41.48 & 30.99 \\
\midrule
\multicolumn{6}{l}{\textit{Tool-Learning Methods}} \\
APIGen-MT ~\cite{prabhakar2025apigen} & 27.66 \dec{3.55} & 17.22 \imp{1.38} & 30.86 \dec{4.57} & 34.18 \dec{7.30} & 27.48 \dec{3.51} \\
TOUCAN ~\cite{xu2025toucan} & 28.75 \dec{2.46} & 19.94 \imp{4.10} & 37.71 \imp{2.28} & 32.89 \dec{8.59} & 29.82 \dec{1.17} \\
Simia ~\cite{li2025simulating} & 21.39 \dec{9.82} & 7.83 \dec{8.01} & 30.86 \dec{4.57} & 31.16 \dec{10.32} & 22.81 \dec{8.18} \\
\midrule
\multicolumn{6}{l}{\textit{TDScaling Implementation (Ours)}} \\
\textbf{TDScaling} & \textbf{33.13} \imp{1.92} & \textbf{23.56} \imp{7.72} & \textbf{39.43} \imp{4.00} & \textbf{43.83} \imp{2.35} & \textbf{34.99} \imp{4.00} \\
\bottomrule
\end{tabular}
\caption{
    Performance on Coding Agent benchmarks. 
    For RebenchT, we report Pass@1 scores using \textbf{O}pen\textbf{H}ands (OH) and \textbf{Qod}er (Qod) agents.
    While baseline methods suffer from negative transfer (indicated by \textcolor{red}{-drop}), particularly in the rigorous Bird benchmark, TDScaling effectively reverses this trend. It is the only method that achieves comprehensive improvements (\textcolor{teal}{+gain}) across all metrics, preserving and enhancing the intrinsic programmatic reasoning.
}

\label{tab:coding_main}
\end{table*}

\begin{table*}[t]
\centering
\small
\setlength{\tabcolsep}{7pt}
\begin{tabular}{lccccccc}
\toprule
\multirow{2}{*}{\textbf{Configuration}} & \multicolumn{2}{c}{\textbf{General Tool-Use}} & \multicolumn{4}{c}{\textbf{Coding \& Agent Capabilities}} & \multirow{2}{*}{\textbf{Average}} \\
\cmidrule(lr){2-3} \cmidrule(lr){4-7}
 & \textbf{BFCL} & \textbf{TAU} & \textbf{RebenchT} & \textbf{RebenchT} & \textbf{CodeCI} & \textbf{Bird} & \\
 & \textit{(Multi-turn)} & \textit{(Avg)} & \textit{OH-p@1} & \textit{Qod-p@1} & \textit{avg@2} & \textit{p@1} & \textit{(Avg)} \\
\midrule
\textbf{TDScaling (Full Model)} & 36.66 & 56.10 & \textbf{33.13} & \textbf{23.56} & 39.43 & \textbf{43.83} & \textbf{38.79} \\
\midrule
\multicolumn{8}{l}{\textit{Ablation Variants}} \\
w/o Cluster Sampling & 34.72 & 54.90 & 29.31 & 20.61 & \textbf{40.57} & 40.54 & 36.78 \\
w/o Global Evolution & 33.64 & 56.00 & 30.30 & 21.85 & \textbf{40.57} & 41.52 & 37.31 \\
w/o Code Tool & \textbf{37.56} & \textbf{56.70} & 28.35 & 21.75 & 38.95 & 41.58 & 37.48 \\
w/o All & 30.25 & 34.05 & 31.30 & 19.00 & 38.29 & 40.12 & 32.17 \\
\bottomrule
\end{tabular}
\caption{
Ablation results. Bold indicates the best score in each column. 
All variants are evaluated on 500 samples to control the API cost of large-scale evaluation. 
Notably, \textbf{w/o Code Tool} scores higher on general tool-use benchmarks (BFCL, TAU) by restricting the action space to API calls, which reduces over-reasoning. 
However, this gain comes with a substantial drop on complex coding tasks. 
TDScaling (Full Model) achieves the best Average score, providing the most robust trade-off between general tool proficiency and programmatic reasoning.
}

\label{tab:ablation_final}
\end{table*}

\subsection{Analysis}
\label{sec:analysis}

\begin{figure}[t]
\centering
\includegraphics[width=0.95\linewidth]{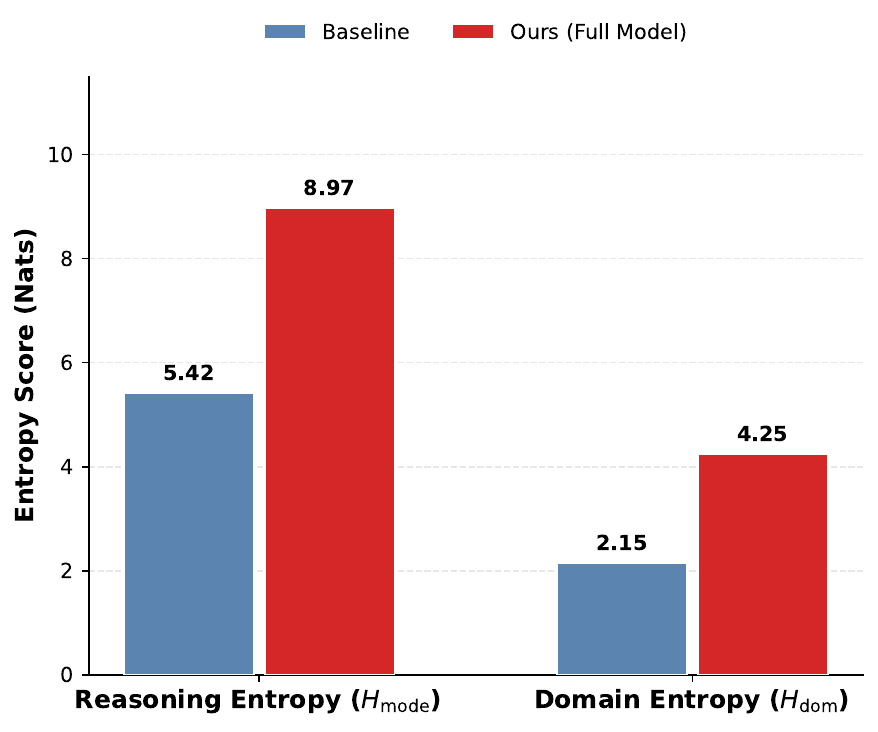}
\caption{
  \textbf{Diversity Analysis.} 
  We quantitatively compare the Reasoning Mode Entropy ($H_{\text{mode}}$) and Domain Entropy ($H_{\text{dom}}$).
  \textbf{Ours} (Red) achieves significantly higher entropy scores ($8.97$ and $4.25$) compared to the \textbf{Baseline} ($5.42$ and $2.15$).
  \textbf{Mechanism:} This performance gap stems from our \textbf{dynamic tagging-and-evolution loop}, where the system autonomously tags generated trajectories with reasoning modes and actively guides subsequent synthesis to not only fill distributional gaps but also \textbf{explore novel, non-prespecified strategies} suitable for complex tool clusters.
}
\label{fig:diversity_analysis}
\end{figure}

\begin{figure}[t]
\centering
\includegraphics[width=\linewidth]{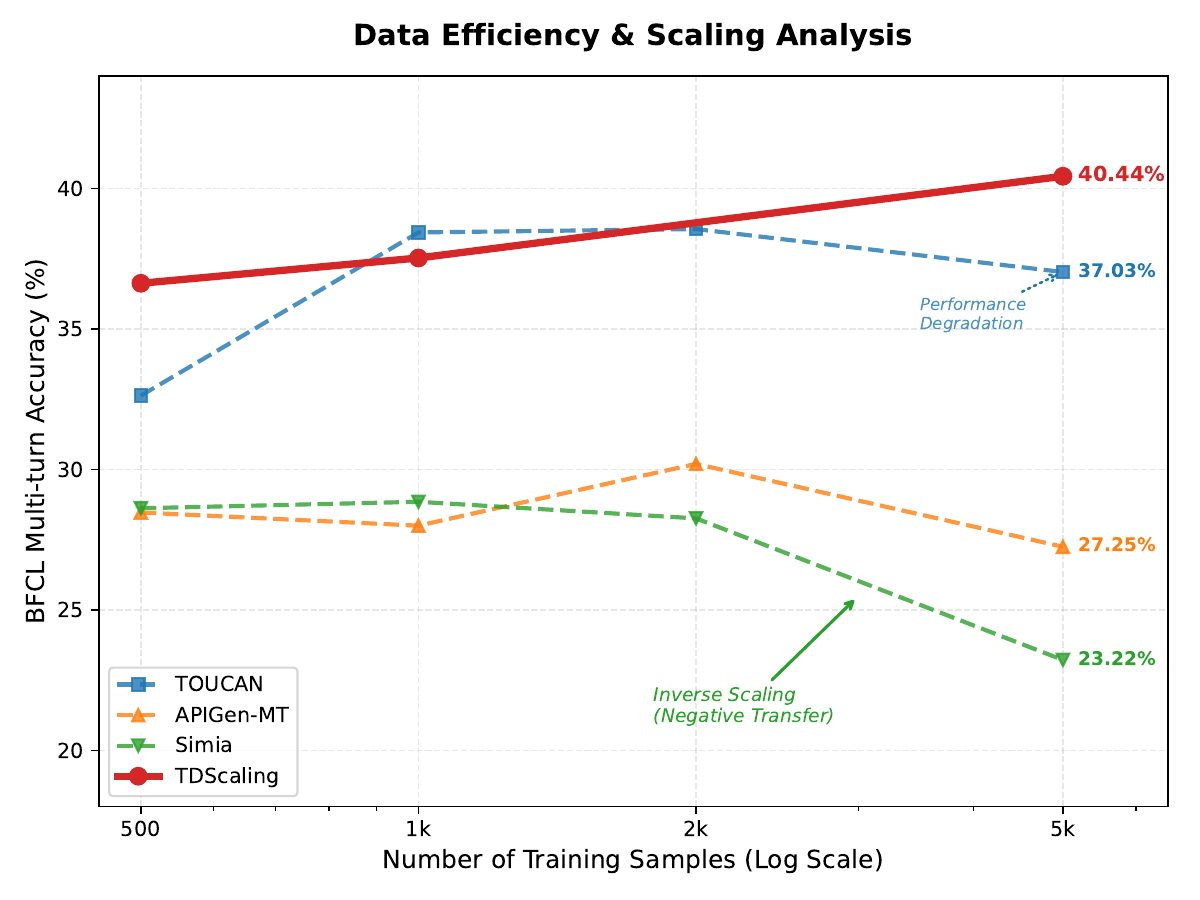}
\caption{
  \textbf{Data Scaling Analysis on BFCL Benchmark.} 
  Our method (Red) achieves strong performance with only 1k samples and reaches \textbf{40.44\%} at 5k samples.
  In contrast, baselines (dashed lines) exhibit \textit{Inverse Scaling}, where performance degrades with more data, indicating overfitting to low-quality, homogeneous patterns.
}
\label{fig:data_scaling}
\end{figure}

\begin{figure*}[h]
\centering
\includegraphics[width=\linewidth]{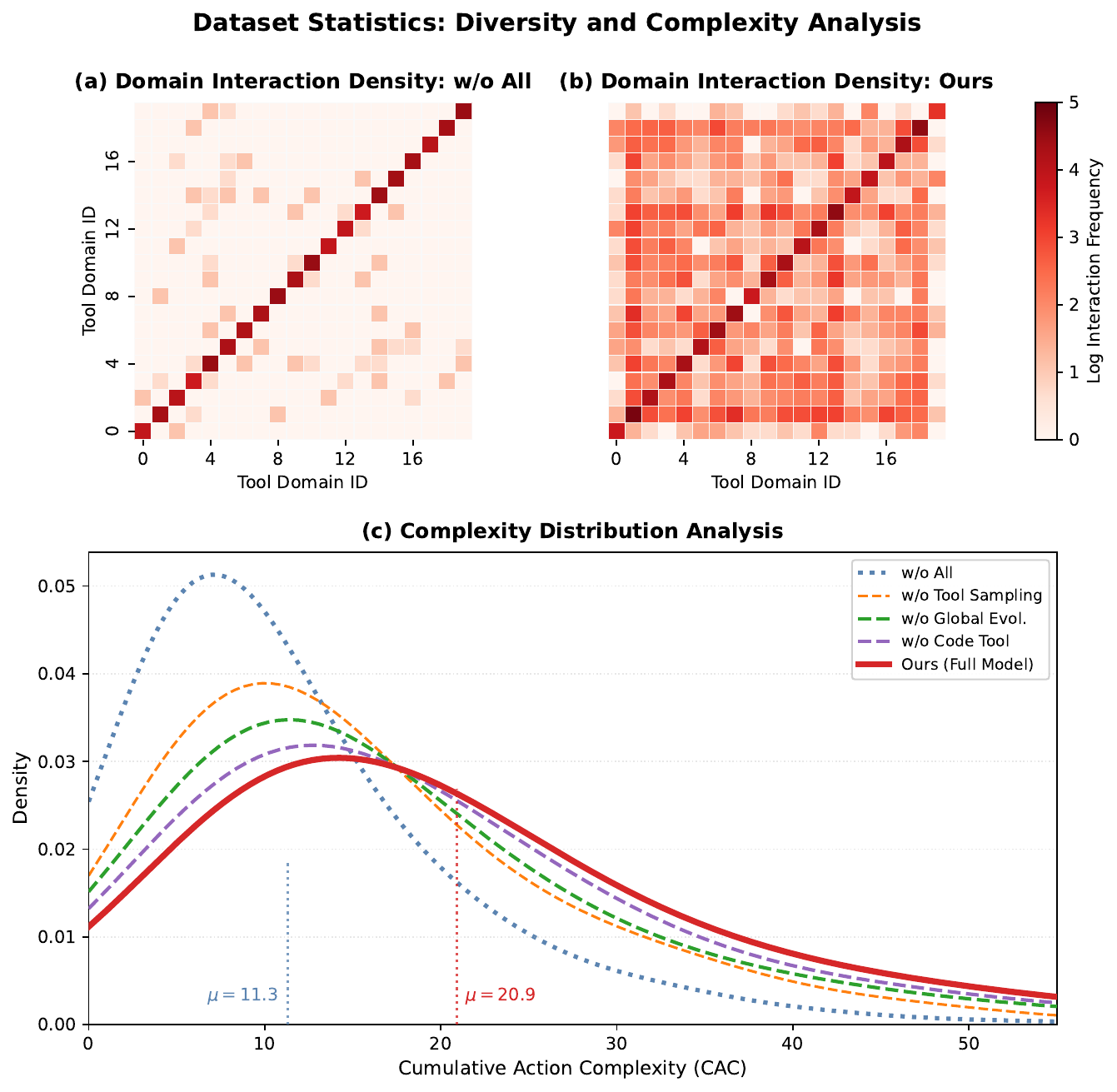}
\caption{
  \textbf{Detailed Dataset Statistics.}
  \textbf{(a-b) Domain Interaction Density:} Ours (b) demonstrates significantly richer cross-domain interactions compared to the sparse \textbf{w/o All} configuration (a).
  \textbf{(c) Complexity Distribution:} The KDE plot confirms that our evolutionary framework generates trajectories with much higher complexity ($\mu=20.9$) than the \textbf{w/o All} baseline ($\mu=11.3$).
}
\label{fig:dataset_stats_appendix}
\end{figure*}

\paragraph{Dataset Quality: Breadth and Depth.}
To quantify dataset quality, we provided formal definitions of the diversity and complexity metrics in Appendix~\ref{app:metric_definitions}.
Our analysis suggested that realizing the full value of synthetic trajectories required improving both breadth (semantic diversity) and depth (structural complexity).
As shown in Figure~\ref{fig:diversity_analysis}, our framework expanded the effective solution space:
\begin{itemize}
    \item \textbf{Breadth (Entropy):} Our data achieved higher Domain Entropy ($H_{\text{dom}}$: 4.25 vs.\ 2.15) and Reasoning Mode Entropy ($H_{\text{mode}}$: 8.97 vs.\ 5.42), indicating broader coverage of domains and reasoning patterns and reduced risk of mode collapse.
    \item \textbf{Depth (Complexity):} As detailed in Appendix~\ref{app:dataset_analysis} and visualized in Figure~\ref{fig:dataset_stats_appendix}(c), our trajectories showed higher Cumulative Action Complexity ($\mu=20.9$) than the \textbf{w/o All} configuration ($\mu=11.3$).
\end{itemize}
The heatmaps in Figure~\ref{fig:dataset_stats_appendix}(a--b) further indicated dense cross-domain interaction patterns that were absent in \textbf{w/o All}. Together, these gains encouraged generalizable behavior rather than memorization of short, repetitive API sequences.

\paragraph{Breaking the Performance Ceiling.}
We examined the effect of training data size in Figure~\ref{fig:data_scaling}.
Several benchmark-targeted synthesis pipelines exhibited inverse scaling, where adding more data degraded performance, consistent with overfitting to noisy and homogeneous patterns in quantity-centric datasets.
In contrast, TDScaling showed positive scaling with strong data efficiency. With only 500 samples, it already reached 36.66\%, establishing a competitive baseline. Scaling to 5,000 samples increased performance to 40.44\%, surpassing the previous SOTA. These results showed that when trajectories maintained sufficient diversity, additional data translated into genuine capability gains, raising the ceiling that constrained quantity-centric synthesis.

\subsection{Ablation Studies}
\label{sec:ablation}

To isolate the effect of each component, we conducted an ablation study on BFCL and the coding benchmarks (Table~\ref{tab:ablation_final}).

\paragraph{Impact of Components.}
The \textit{w/o All} variant achieved the lowest performance, indicating that naive synthesis did not induce sufficient reasoning depth.
Removing Global Evolution reduced performance on the more complex benchmarks, consistent with diminished trajectory complexity.
Removing the Code Tool exposed a clear trade-off: BFCL remained competitive (37.56\%), but coding performance dropped (BIRD: 41.58\% vs.\ 43.83\%). This pattern suggested that the Code Tool functioned as a regularizer, encouraging precise algorithmic reasoning and code generation and thereby mitigating catastrophic forgetting.

\section{Conclusion}

This study redefines data scaling for code agents: the limiting factor for generalization is not the volume of trajectories, but the \textbf{trajectory diversity} available. We demonstrate that trajectory diversity serves as a first-class optimization target, providing a principled route to expand the effective solution space and raise performance ceilings under fixed data budgets. By shifting focus from raw quantity to structural coverage, TDScaling overcomes the diminishing returns and inverse scaling observed in homogeneous synthetic datasets. Furthermore, we identify that tool proficiency should not be optimized in isolation from core programming competence. Pure tool tuning risks eroding intrinsic coding skills—negative transfer—whereas coupling API interactions with programmatic reasoning via a sandboxed code tool acts as a stabilizer. This synergy preserves the model's pre-training priors while strengthening its ability to handle complex, logical workflows, turning a common trade-off into a complementary gain.

Ultimately, TDScaling establishes that synthetic data pipelines must be \emph{measured} and \emph{directed} rather than randomly generated. By actively steering synthesis toward gaps in domain entropy and complexity, we offer an actionable framework for resource-efficient training that aligns with the evolving modularity of the Model Context Protocol. We release our framework and dataset to encourage more refined diversity metrics, and to facilitate further research into how diversity and programmatic grounding jointly determine the robustness of next-generation code agents.

\section*{Limitations}

While TDScaling improves data efficiency, it has two main limitations. First, synthesis is compute- and cost-intensive. Achieving logically consistent, high-diversity trajectories requires a multi-agent pipeline that depends on high-capability teacher models for planning, execution, and verification. As a result, the per-trajectory API cost and latency are higher than lightweight baselines such as simple rejection sampling. In this work, we intentionally traded generation speed for higher quality density.

Second, our current interaction scope is limited to text-based tool calls and Python execution. Many real-world agents must also operate over graphical user interfaces and visual web content, where observations are multi-modal and state transitions can be harder to represent. Extending our diversity and complexity signals to multi-modal settings, and validating whether the same diversity-scaling behavior holds, is an important direction for future work.

\section*{Ethical Considerations}  
This work introduced TDScaling, a data synthesis framework designed for software engineering tasks and API tool-use scenarios. All datasets and experiments relied on publicly available technical documentation and open-source benchmarks. We complied with the usage policies and licenses of the base models (e.g., Qwen) and all benchmark datasets used in this study.

Because our study relied exclusively on synthetic data generated by LLMs, it did not involve human subjects, crowdsourcing, or personally identifiable information (PII). We nonetheless acknowledged known risks of code LLMs, including the potential to generate insecure or malicious code. To reduce this risk, our synthesis pipeline incorporated strict quality filtering and sandboxed execution to improve the safety and reliability of generated trajectories. Beyond these established concerns, we did not identify additional societal harms specific to our setting relative to the broader literature on general-purpose code LLMs.


\bibliography{custom}

\appendix
\appendix

\section{Detailed Dataset Analysis}
\label{app:dataset_analysis}

To understand the source of our model's performance improvements, we conduct a fine-grained analysis of the synthesized dataset (\textbf{TDScaling}) compared to the Baseline.

\paragraph{Domain Interaction Density.}
Figures~\ref{fig:dataset_stats_appendix}(a) and (b) visualize tool co-occurrence patterns. 
Specifically, both the x-axis and y-axis represent unique Tool Domain IDs. The heatmap is constructed as a co-occurrence matrix where the value at coordinate $(i, j)$ represents the frequency of trajectories that involve tools from both Domain $i$ and Domain $j$. For instance, if a single trajectory invokes a tool from Domain 0 and a tool from Domain 1, the interaction count at $(0, 1)$ is incremented.
The \textbf{Baseline (a)} exhibits a strong diagonal pattern, indicating that random sampling predominantly generates trajectories confined to single domains (such as only using Search tools).
In contrast, \textbf{Ours (b)} displays a dense network of off-diagonal activations. High-intensity blocks (dark red) appear in cross-domain regions, representing frequent collaborative usage between distinct tool categories (such as Code Tool interacting with Data Analysis tools). This confirms that our Global Memory mechanism successfully identifies and fills gaps in tool combinations.

\paragraph{Complexity Distribution.}
Figure~\ref{fig:dataset_stats_appendix}(c) presents the Cumulative Action Complexity (CAC) distribution via Kernel Density Estimation (KDE).
\begin{itemize}
    \item The \textbf{Baseline} (Blue dotted line) is heavily skewed towards the lower spectrum ($\mu=11.3$), lacking the long-tail characteristic required for agentic tasks.
    \item \textbf{Ours} (Red solid line) shifts the probability mass significantly to the right ($\mu=20.9$), with a heavy tail indicating the presence of long-horizon, multi-step reasoning chains.
\end{itemize}

\section{Metric Definitions}
\label{app:metric_definitions}

We provide the formal mathematical definitions for the complexity and diversity metrics used in our evaluation.

\subsection{Cumulative Action Complexity (CAC)}
To quantify hierarchical complexity, we analyze the dependency depth of each parameter $p$ in a tool call $\theta$. We classify dependencies into three levels:
\begin{itemize}
    \item \textbf{Instruction-Grounded ($\omega_1=1.0$)}: Derived directly from the user query or static constants.
    \item \textbf{Local-Context ($\omega_2=1.1$)}: Depends on the immediate previous turn.
    \item \textbf{Global-Context ($\omega_3=1.2$)}: Requires multi-turn retrieval or synthesis from earlier states.
\end{itemize}
The complexity of a single tool call is determined by the bottleneck principle:
\begin{equation}
\mathcal{C}_{\text{depth}}(\theta \mid \mathcal{H}) = \max_{p \in \theta} \omega_{y(p)}
\end{equation}
The total CAC is the sum of step-wise complexities plus switching costs ($\delta=0.2$) for cross-domain transitions.

\subsection{Diversity Metrics (Entropy)}
To rigorously quantify dataset quality, we define two entropy-based metrics.

\paragraph{Reasoning Mode Entropy ($H_{\text{mode}}$).}
We categorize reasoning traces into discrete modes $\mathcal{M}$ (such as Direct Execution, Error Correction, Multi-step Planning, Reflection). The entropy is calculated as:
\begin{equation}
H_{\text{mode}} = - \sum_{m \in \mathcal{M}} p(m) \log p(m)
\end{equation}
A higher $H_{\text{mode}}$ indicates a broader spectrum of cognitive behaviors beyond simple execution.

\paragraph{Domain Entropy ($H_{\text{dom}}$).}
To capture the semantic breadth of the synthesized dataset, we compute entropy at the \textbf{Business Cluster} level, consistent with the formulation in Section~\ref{sec:metric_definitions}:
\begin{equation}
H_{\text{dom}} = - \sum_{B_k \in \mathcal{B}} p(B_k) \log p(B_k)
\end{equation}
where $p(B_k)$ is the normalized frequency of trajectories belonging to cluster $B_k$. High entropy implies a uniform distribution across diverse service domains, avoiding over-concentration on common tools.

\section{Implementation Details}
\label{app:implementation_details}

\subsection{Tool-Space Construction}
We construct a composite feature text $T_{\text{feat}}(t)$ for each tool and encode it using \textbf{Qwen-Embedding-0.6B}. We employ a two-level K-Means algorithm:
\begin{enumerate}
    \item \textbf{Latent Domain Partitioning:} Partitions tools into broad domains ($N_{\text{dom}} = 10$).
    \item \textbf{Functional Class Abstraction:} Further clusters tools into fine-grained classes ($N_{\text{cls}} = 5$).
\end{enumerate}

\paragraph{Illustrative Example of Selection.}
Consider three clusters covering classes $\{1,2,3\}$ ($B_1$), $\{4\}$ ($B_2$), and $\{2,5\}$ ($B_3$). Given a budget of 2:
\begin{enumerate}
    \item \textbf{Step 1:} Select $B_1$ (Adds 3 classes: 1,2,3).
    \item \textbf{Step 2:} $B_2$ adds class 4 (Gain=1). $B_3$ adds class 5 (Gain=1, since 2 is covered).
    \item \textbf{Outcome:} Tie-breaking selects $B_3$. Final set $\{B_1, B_3\}$ maximizes diversity while retaining intra-cluster logic (keeping the overlapping Class 2).
\end{enumerate}

\subsection{Training Setup}
We utilize \texttt{Megatron-LM} for distributed training on a cluster of compute nodes, where each node is equipped with 8 $\times$ 80GB GPUs.
\begin{itemize}
    \item \textbf{Base Model:} Qwen3-Coder-30B-A3B-Instruct.
    \item \textbf{Hyperparameters:} Global Batch Size 16, Learning Rate 1e-5 (Cosine decay, 50 warmup steps).
    \item \textbf{Context:} Sequence Length 65,536 tokens.
    \item \textbf{Optimization:} BF16 precision, Flash Attention v2, Gradient Checkpointing enabled.
\end{itemize}

\section{System Prompts and Case Study}
\label{app:prompts_and_case}

We provide the qualitative materials to reproduce our method. Section~\ref{app:prompts_system} details the consolidated system instructions, and Section~\ref{app:case_study} presents a concrete execution trajectory.

\subsection{Detailed System Prompts}
\label{app:prompts_system}

To facilitate reproducibility, we consolidate the core system instructions into three categories: Foundation \& Blueprinting (Figure~\ref{fig:prompt_foundation}), Interactive Role-Playing (Figure~\ref{fig:prompt_roleplay}), and Environment \& Evaluation (Figure~\ref{fig:prompt_environment}).
Crucially, the \textbf{BlueprintAgent} component receives dynamic inputs from the Global Memory to steer the evolution of task complexity.

\begin{figure*}[h]
\begin{promptbox}{System Instruction: Shared Anti-Hallucination Constraints}
\textbf{STRICT ANTI-HALLUCINATION PROTOCOL:}
1. \textbf{Verified Numerical Data Only:} You are prohibited from providing specific numerical values (costs, prices, fees, measurements) unless you have obtained them from a tool call result in the current conversation turn.
2. \textbf{No Synthetic Completion:} Do not generate phrases like "Let me synthesize a final answer" unless citing specific evidence retrieved from tools.
3. \textbf{Tool-First Logic:}
   - Step 1: Call the appropriate tool (API or Code Interpreter).
   - Step 2: Wait for the tool execution result.
   - Step 3: Report numbers based strictly on the output.
   - If a tool fails: State "The tool did not return data." Do not fabricate values.

\textbf{Consistency Enforcement:}
- \textbf{Status Consistency:} Do not claim a task is complete if dependency steps remain unexecuted.
\end{promptbox}

\vspace{0.2cm}

\begin{promptbox}{BlueprintAgent: Scenario Blueprinting \& Persona Generation}
\textbf{Objective:} Design high-quality, natural, and realistic \textbf{Scenario Blueprints} based on the provided tool definitions.

\textbf{DYNAMIC STRATEGY INPUTS (From Global Memory):}
\textit{\{strategy\_profile\_placeholder\}}
(Examples of injected directives: 
 "Target: Increase 'Error Recovery' samples if tools support complex parameters.", 
 "Target: 'Multi-step Planning' preferred; check data dependencies first.",
 "Entropy Goal: Avoid 'Direct Execution'; explore novel usage patterns.")

\textbf{STRATEGY ADAPTATION \& FEASIBILITY CHECK (CRITICAL):}
You are the \textbf{Judge} of the strategy's viability.
1. \textbf{Assess Tool Fit:} Look at the provided tool definitions. Does the Global Strategy fit these specific tools?
   - \textit{Example:} If Global asks for "Error Correction" but the tool is a simple \texttt{get\_time()} API, this is a \textbf{Mismatch}.
2. \textbf{Decision Logic:}
   - \textbf{If Matched:} Design a scenario that explicitly forces the requested strategy (e.g., provide invalid inputs to trigger recovery).
   - \textbf{If Mismatched:} \textbf{Override the instruction.} Do not force a bad fit. Instead, brainstorm a \textbf{novel interaction pattern} that uniquely exploits the features of \textit{this} tool cluster.

\textbf{BRAINSTORMING REQUIREMENTS:}
1. \textbf{User Goal:} Design a goal that is SPECIFIC, MULTI-FACETED, and CHALLENGING. 
   - \textit{Example of Depth:} "Troubleshoot a database latency spike: query system logs for error spikes, check CPU utilization metrics, and if usage is high, fetch the slow query log to identify the bottleneck."
2. \textbf{Complexity Alignment:} 
   - Simple tasks: 2-4 turns; Complex tasks: 7-12 turns.

\textbf{PERSONA GENERATION:}
\begin{itemize}[leftmargin=*, nosep]
\item \textbf{User Persona:} Specific identity (such as "Hurried Data Analyst") with traits (such as "impatient and results-focused").
\item \textbf{Assistant Persona:} Role (such as "Efficient Expert") with balanced verbosity.
\end{itemize}

\textbf{Tool Evolution Plan:}
Identify \texttt{missing\_tools} needed for the workflow and propose specific functional requirements.
\end{promptbox}
\caption{
    \textbf{Foundation Constraints and Blueprinting.} 
    Top: The shared protocols injected into all agents to ensure factual grounding. 
    Bottom: The \textbf{BlueprintAgent} prompt, which integrates \textbf{Dynamic Strategy Inputs} to actively steer synthesis toward under-explored complexities.
}
\label{fig:prompt_foundation}
\end{figure*}

\begin{figure*}[h]
\begin{promptbox}{UserAgent System Prompt}
You are a realistic human user. 

\textbf{Critical Thinking:}
- \textbf{Skepticism:} If the assistant claims a fact without evidence, query it.
- \textbf{Spot Contradictions:} If the assistant contradicts the tool output, point it out.
- \textbf{Demand Evidence:} If the assistant is vague, ask "What exactly did you find?"

\textbf{Stylistic Constraints:}
- \textbf{Natural Language:} Be casual, varied, and allow for minor human imperfections.
- \textbf{Avoid Formulaic Openers:} Do not start every turn with "Great", "Perfect", or "Okay".
\end{promptbox}

\vspace{0.2cm}

\begin{promptbox}{AssistantAgent System Prompt}
You are a helpful, rigorous, and honest AI assistant.

\textbf{CONVERSATION STYLE CONSTRAINTS:}
\begin{itemize}[leftmargin=*, nosep]
\item \textbf{Restricted Openers:} Avoid repetitive usage of "Great", "Perfect", "Excellent".
\item \textbf{Conciseness:} Do not use robotic acknowledgments like "I understand" or "Noted" unless necessary for clarity.
\end{itemize}

\textbf{Interaction Logic:}
- \textbf{Tool-Only Turns:} If executing a tool, do not generate simultaneous chat content unless necessary for reasoning traces.
- \textbf{Error Handling:} If a tool fails, admit the failure clearly. Do not hallucinate a successful outcome.
\end{promptbox}
\caption{\textbf{Interactive Role-Playing Prompts.} We instruct the UserAgent to be skeptical and natural, while the AssistantAgent is constrained to be rigorous and concise, preventing the "yes-man" bias common in synthetic data.}
\label{fig:prompt_roleplay}
\end{figure*}

\begin{figure*}[h]
\begin{promptbox}{ObservationAgent System Prompt}
\textbf{Role:} World-class API response simulator. Generate production-grade JSON responses.

\textbf{Simulation Rules:}
1. \textbf{Format Strictness:} Return ONLY valid JSON. No markdown, no explanations.
2. \textbf{Realism:} Use realistic values (such as logical travel times, non-zero prices).

\textbf{Consistency \& Schema Locking:}
\begin{itemize}[leftmargin=*, nosep]
    \item \textbf{Temporal Consistency:} Timestamps must follow a logical sequence across turns.
    \item \textbf{Entity Consistency:} IDs and names for the same entity must not change.
    \item \textbf{Dynamic Schema Locking:} Once a tool's output structure is generated in Turn $T$, strict adherence to this structure is enforced in Turns $T+1 \dots N$.
\end{itemize}
\end{promptbox}

\vspace{0.2cm}

\begin{promptbox}{QualityAgent Evaluation Rubric}
Evaluate this conversation trajectory for training data suitability.

\textbf{EVALUATION DIMENSIONS (Score 0-10):}
1. \textbf{Realism \& Fluidity:} Natural human patterns, appropriate hesitation.
2. \textbf{Tool Usage Intelligence:} Strategic selection, meaningful chaining.
3. \textbf{Anti-Hallucination (CRITICAL):} 
   - Did the assistant use "synthesize" without evidence?
   - Did it provide facts NOT backed by tools?
   - \textit{If ANY hallucination is detected, mark suitable\_for\_training = False.}
4. \textbf{Goal Achievement:} Did the conversation achieve the intended user goal?
\textbf{CATEGORIZATION:}
\textbf{Reasoning Mode Label:} Classify the interaction strategy into one tag (e.g., \textit{Direct Execution}, \textit{Error Correction}, \textit{Multi-step Planning}, \textit{Reflection}). You may define a new tag if the behavior is unique.
\textbf{AUTOMATIC REJECTION CRITERIA:}
- Specific numbers/facts appearing without tool backing.
- Assistant "predicting" values before execution.
- Tool results contradicting assistant's pre-tool statements.
\end{promptbox}
\caption{\textbf{Environment Simulation and Quality Control.} Top: The ObservationAgent employs Dynamic Schema Locking to prevent structural hallucinations. Bottom: The QualityAgent filters trajectories based on rigorous realism and consistency checks.}
\label{fig:prompt_environment}
\end{figure*}

\subsection{Case Study}
\label{app:case_study}
\begin{figure*}[h]
\centering
\includegraphics[width=\linewidth]{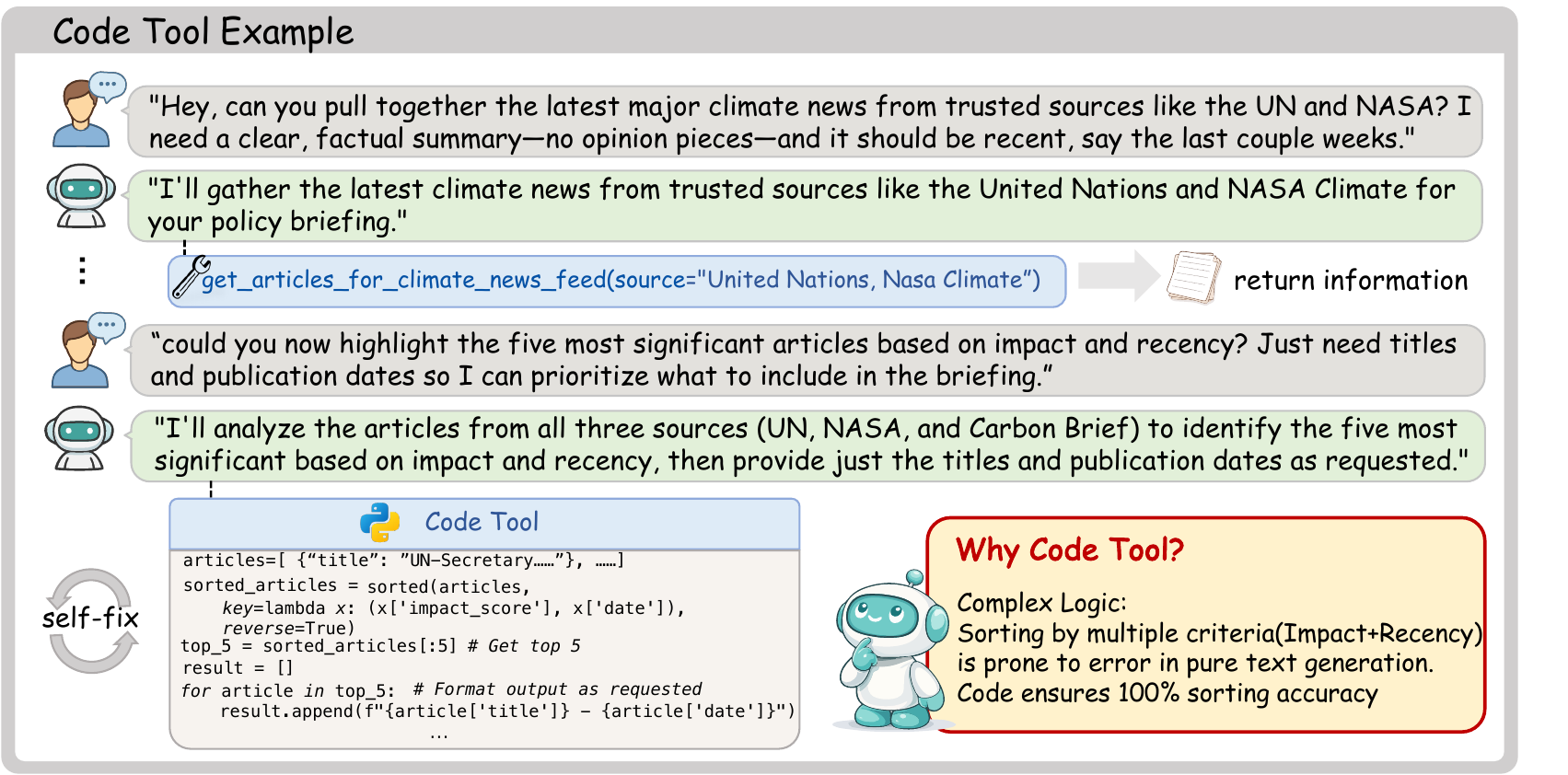}
\caption{\textbf{Code Tool Case.}}
\label{fig:code_tool_appendix}
\end{figure*}

To further elucidate the motivation behind integrating the Code Tool into our data synthesis framework, we present a concrete interaction scenario in Figure~\ref{fig:code_tool_appendix}. Relying solely on an LLM's internal Chain-of-Thought to sort and filter a large JSON object is computationally expensive and prone to calculation hallucinations, particularly when the context window is filled with structured data. By introducing the Code Tool, the agent can offload this computational burden to a deterministic Python interpreter. As shown in the generated snippet, the agent writes a lambda function to sort the dictionary precisely. This paradigm significantly enhances robustness in tasks involving arithmetic, sorting, or complex logic constraints.

\begin{figure*}[h]
\centering
\includegraphics[width=\linewidth]{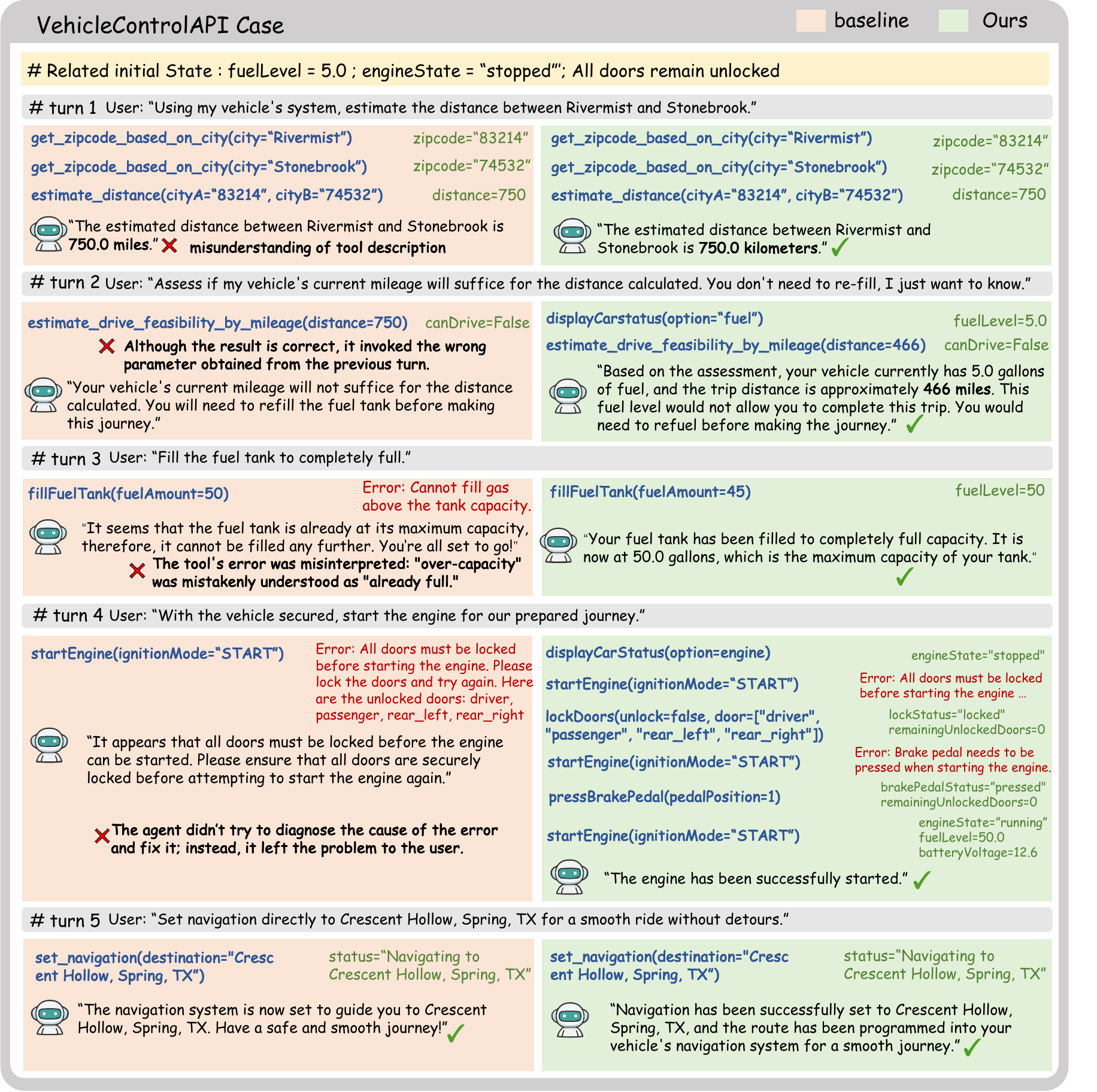}
\caption{\textbf{Annotated Case Study.} A generated trajectory demonstrating the agent's capability in handling parameter dependencies and error signals (highlighted in boxes).}
\label{fig:case_study_appendix}
\end{figure*}

Figure~\ref{fig:case_study_appendix} presents a complete trajectory involving the \textit{VehicleControlAPI}. This case demonstrates the model's ability to:
(1) Handle complex multi-turn interactions with precise parameter usage.
(2) Correctly identify tool failures and perform error recovery without user intervention.
(3) Maintain context consistency across cross-domain tool calls.

\end{document}